# Toward autocorrection of chemical process flowsheets using large language models

Lukas Schulze Balhorn,[a] Marc Caballero,[b] and Artur M. Schweidtmann[a]

[a]*Process Intelligence Research Group, Department of Chemical Engineering, Delft University of Technology, Van der Maasweg 9, Delft 2629 HZ, The Netherlands*
[b]*Department of Chemical Engineering, Delft University of Technology, Van der Maasweg 9, Delft 2629 HZ, The Netherlands*
a.schweidtmann@tudelft.nl

## Abstract

The process engineering domain widely uses Process Flow Diagrams (PFDs) and Process and Instrumentation Diagrams (P&IDs) to represent process flows and equipment configurations. However, the P&IDs and PFDs, hereafter called flowsheets, can contain errors causing safety hazards, inefficient operation, and unnecessary expenses. Correcting and verifying flowsheets is a tedious, manual process. We propose a novel generative AI methodology for automatically identifying errors in flowsheets and suggesting corrections to the user, i.e., autocorrecting flowsheets. Inspired by the breakthrough of Large Language Models (LLMs) for grammatical autocorrection of human language, we investigate LLMs for the autocorrection of flowsheets. The input to the model is a potentially erroneous flowsheet and the output of the model are suggestions for a corrected flowsheet. We train our autocorrection model on a synthetic dataset in a supervised manner. The model achieves a top-1 accuracy of 80% and a top-5 accuracy of 84% on an independent test dataset of synthetically generated flowsheets. The results suggest that the model can learn to autocorrect the synthetic flowsheets. We envision that flowsheet autocorrection will become a useful tool for chemical engineers.

## 1. Introduction

The process engineering domain widely uses Process Flow Diagrams (PFDs) and Process and Instrumentation Diagrams (P&IDs) to represent process flows and equipment configurations. However, the P&IDs and PFDs, hereafter called flowsheets, can contain errors like missing or misplaced components, incorrect signal or stream connections, or even missing or misplaced subsystems. These errors can cause significant safety hazards, delays in development, inefficient operation, and unnecessary expenses. Hence, identifying and correcting errors in flowsheets is important but currently a tedious, manual process.

In the scientific literature, there are a few initial publications on the error detection and corrections of process flowsheets. These works are based on two main concepts: (i) rule-based approaches and (ii) machine learning-based (ML) approaches. Rule-based approaches encode engineering rules for common errors, e.g., as graph patterns, and then detect and correct the errors, e.g., through graph manipulations. For example, Bayer and Sinha (2019) and Bayer et al. (2022) define and apply graph patterns that contain erroneous patterns and their corrections. Similarly, Shin et al. (2023) encode graph manipulations to correct the setup of compression devices. The definition of graph patterns and manipulations makes the approach understandable for users. However, this approach relies on graph isomorphism to find the graph patterns in the P&ID graph which is computationally expensive. In addition, rule-based approaches are limited to hard-coded rules which are difficult to develop, maintain, and extend.

Besides rule-based approaches, a few initial ML-based approaches have recently been proposed for error correction. Dzhusupova et al. (2022) detect four engineering error patterns directly on P&ID images using object detection. For instance, they detect configurations where the symbol of a spectacle blind is placed right next to the symbol of a butterfly valve which might cause the blind and the disk to clash if the butterfly valve is opened. A drawback of this approach is its low generalizability to new drawing styles and that it can only detect errors involving items with close visual proximity. Using an ML-based method, Rica et al. (2022) detect anomalies in process graphs. In particular, they create a simple graph embedding by counting neighboring unit operations. This embedding is the input to a multi-layer perceptron (MLP) that predicts the component type of the component of interest. In case there is a mismatch between the predicted and the actual component type, an anomaly is indicated. Furthermore, Shin et al. (2023) train a support vector machine to classify common subgraphs from process graphs as correct or anomaly. Mizanur Rahman et al. (2021) utilize a node degree comparison and a graph convolutional network (GCN) to detect anomalies on a process graph via binary node classification. Recently, Oeing et al. (2022, 2023) also use a GCN to detect inconsistencies in the process design. For each node in a process graph, they predict the component type and compare it with the actual component type (similar to Rica et al., 2022). The first works on ML for autocorrections are promising but there are still a few conceptual limitations. Most ML approaches only check one component individually at a time. Therefore, they cannot detect missing components, incorrect connections, or engineering mistakes involving two or more components. In addition, the computational costs of analyzing one component at a time scales linearly with the number of components in the P&ID which could result in long runtimes for large P&IDs.

For grammatical error correction in human language, large language models (LLMs) proved to be successful (Alikaniotis et al., 2019). In the context of chemical engineering flowsheets, we recently demonstrated that LLMs can autocomplete flowsheets (Vogel et al., 2023). This technology represents flowsheets as strings using the SFILES 2.0 notation (d'Anterroches, 2005; Vogel et al., 2023a) and uses transformer language models to autocomplete flowsheets. Furthermore, we recently formulated the development of P&IDs as a machine translation problem where flowsheets without a control structure are translated to flowsheets with a control structure (Hirtreiter et al., 2023).

In this study, we propose to formulate the autocorrection of flowsheets as a machine translation problem where potentially erroneous flowsheets are translated to correct flowsheets. In particular, we train a transformer language model with flowsheet pairs where the input is a potentially erroneous flowsheet and the output is a corrected flowsheet. Thereby, the model can learn complex error patterns from data and errors in the context of complete flowsheets. For training, we generate synthetic flowsheet pairs with predefined error patterns.

## 2. Autocorrection methodology using transformer models

Our proposed autocorrection model is based on the sequence-to-sequence transformer model using the T5-small transformer model (Raffel et al., 2020). The input to the model is a flowsheet encoded as a string in the SFILES 2.0 notation (Vogel et al., 2023a). The model then generates a new flowsheet which is a corrected version of the input. We can derive suggestions for correction from the new flowsheet by comparing the model input with the model output. As the model generates complete flowsheets, our approach is not limited to modifications of single components.

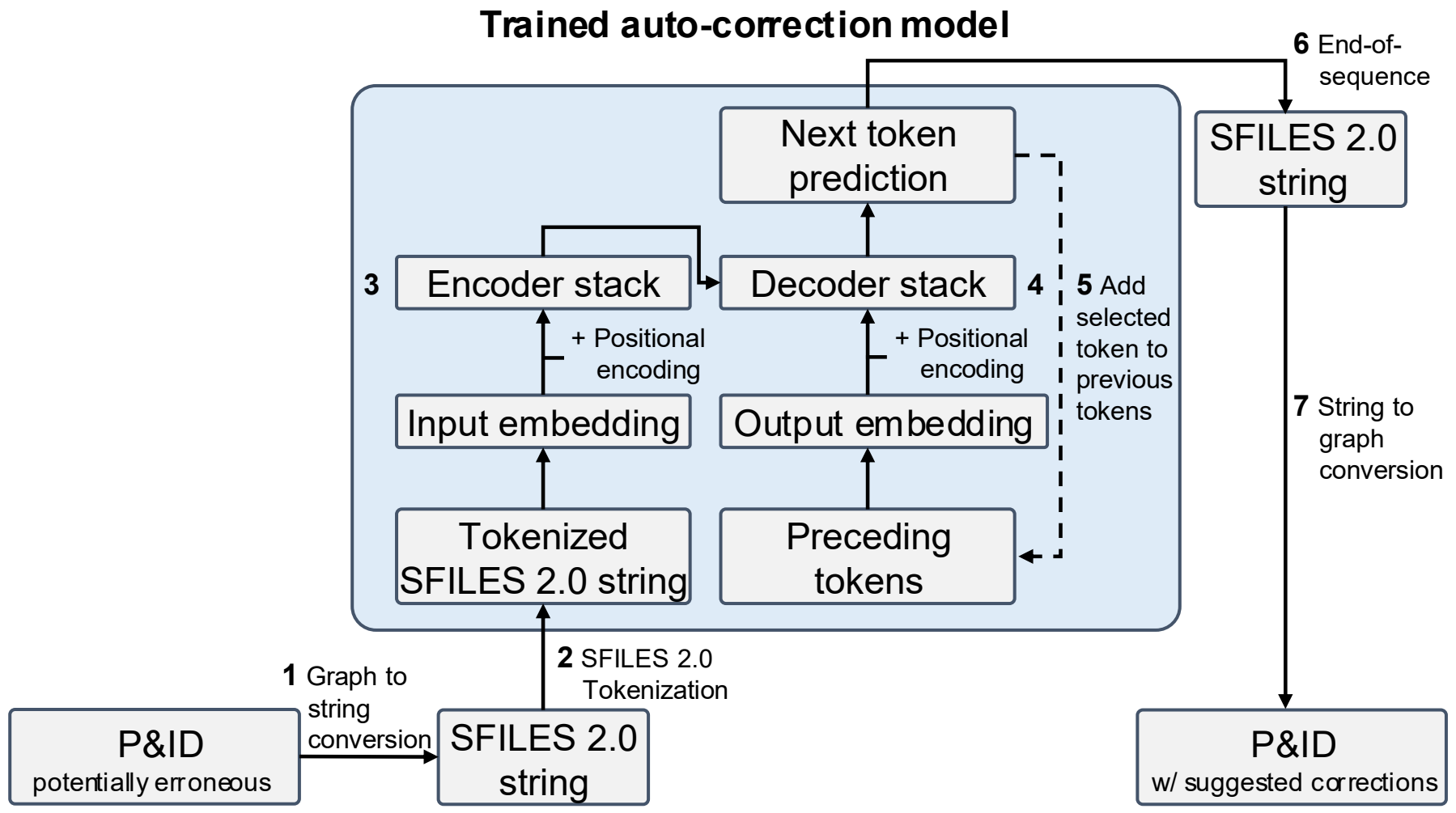

Figure 1. Overview of the autocorrection methodology with the transformer model.

Figure 1 provides an overview of the methodology. The methodology starts with a potentially erroneous flowsheet graph. In Step 1, the flowsheet graph is serialized to a string using the SFILES 2.0 notation (Vogel et al., 2023a). No topological information is lost in this conversion. Then, the SFILES is broken into tokens (Step 2). A token is a small text piece defined in the SFILES vocabulary that represents, for instance, unit operations, raw materials, products, a start-of-sequence (SOS) token, or an end-of-sequence (EOS) token (c.f. Vogel et al., 2023). The SFILES 2.0 vocabulary contains 53 tokens in total. In Step 3, the tokens are mapped to a vector space and their position is encoded. For all vectors, the encoder stack calculates a numerical representation, i.e., the embedding, of the flowsheet using the attention mechanism (Vaswani et al., 2017). The attention mechanism processes all token vectors in parallel which is an advantage over previous recurrent network architectures. Then, Steps 4 and 5 iteratively predict the output tokens. In Step 4, the decoder stack takes two different inputs. One is the numerical embedding from the encoder stack. The other one is the embedding of the preceding output tokens. In Step 5, the decoder stack predicts a likelihood for every token in the vocabulary. The next token is selected using beam search based on the maximum likelihood. For example, in the first iteration, the preceding token is only the SOS token. The decoder stack then predicts the raw material token <raw> as the next token. In the second iteration, the preceding tokens are SOS and <raw>. Steps 4 and 5 are repeated until the EOS token is selected. In Step 6, the sequence of tokens is translated back into an SFILES string. Finally, the SFILES string is converted to a graph in Step 7, thereby generating the corrected flowsheet suggestions.

## 3. Dataset

The autocorrection model is a supervised machine-learning approach. Therefore, it requires source-target data. In our case, the training data contains pairs of flowsheets where one flowsheet is erroneous, the source flowsheet, and the other one is the corresponding correct flowsheet, the target flowsheet. In this proof of concept, we generate such pairs synthetically by using common error patterns in flowsheets. In the future, we plan to train our algorithm on industrial flowsheets.

We define 27 patterns commonly occuring in flowsheets (Towler and Sinnott, 2021; Svrcek et al., 2014; Shinskey, 1979). Each pattern focuses on one of the following unit operations: Mixer, heat exchanger, pump, compressor, storage, adding reactant, reactor, or column. For each pattern, we define up to nine erroneous versions. We aggregate different patterns to complete flowsheets using a Monte Carlo graph generation approach. For a more detailed description of the synthetic flowsheet generation, we refer to Hirtreiter et al. (2023). For 40% of the flowsheet pairs, we include erroneous patterns for the input flowsheet, while the remaining 60% contain only correct flowsheet pairs (i.e., source and target are the same correct flowsheet). This idea is inspired by error correction of human language (Yang et al., 2022) and avoids that the model expects all flowsheets to be erroneous. The optimal ratio of erroneous input flowsheets depends on the model application and is subject to future research. We generate a dataset of 500,000 unique synthetic flowsheet pairs, of which we use 80% to train the autocorrection model, 10% as a validation dataset to monitor the training and detect potential overfitting, and 10% as an independent test dataset. All datasets have a similar ratio of correct and erroneous flowsheet pairs.

## 4. Results and discussion

To optimize the hyperparameters of the model, we perform a grid search. The grid search includes the following hyperparameter ranges with the optimum highlighted in bold: (i) The dimension of a vector that embeds a token (**128**, 256, 512), (ii) the number of layers in the encoder and decoder stack (2, **4**, 6), (iii) the learning rate (1e-4, **5e-4**, 1e-3), and (iv) the batch size (8, 16, **32**). The final model has 7.9 million trainable parameters.

The (top-1) accuracy for predicting the target flowsheet is 80.1% on the independent test dataset. If we consider the five beams with the highest maximum likelihood in the beam search instead of only one, we get five different predictions from the autocorrection model. Considering these five predictions the autocorrection model reaches a (top-5) accuracy of 83.6%. In our test dataset, we assume that a unique correct solution exists which is not a general case. There might exist other correct solutions.

Figure 2 shows the model predictions for an illustrative case study from the independent test set. Note that the test data consist of synthetic flowsheets that do not represent an existing process. However, they fulfill the purpose of illustrating the model's potential. The case study process includes a reaction with a gaseous product which is separated with a distillation column into two product streams. The input flowsheet of the case study contains two potential design errors. Firstly, a pressure controller is missing for the reactor. Pressure can build up in the reactor from the gaseous product of the reaction. The pressure controller is crucial to regulate the mass flow to the downstream system. Secondly, a temperature controller is missing for the heat exchanger before the column. Therefore, the flow controller of the service stream valve does not get feedback from the product stream temperature. The autocorrection model detects both errors and corrects them in the output flowsheet as illustrated in Figure 2. The model adds a temperature controller to the heat exchange and connects it to the product stream and the flow controller. In addition, the model adds a pressure controller to the reactor to control the pressure relief. The suggested corrections from the model correspond to the target flowsheet (i.e., the corrected flowsheet) and is thus considered a correct model prediction. Notably, other feasible corrections of the input flowsheet may exist.

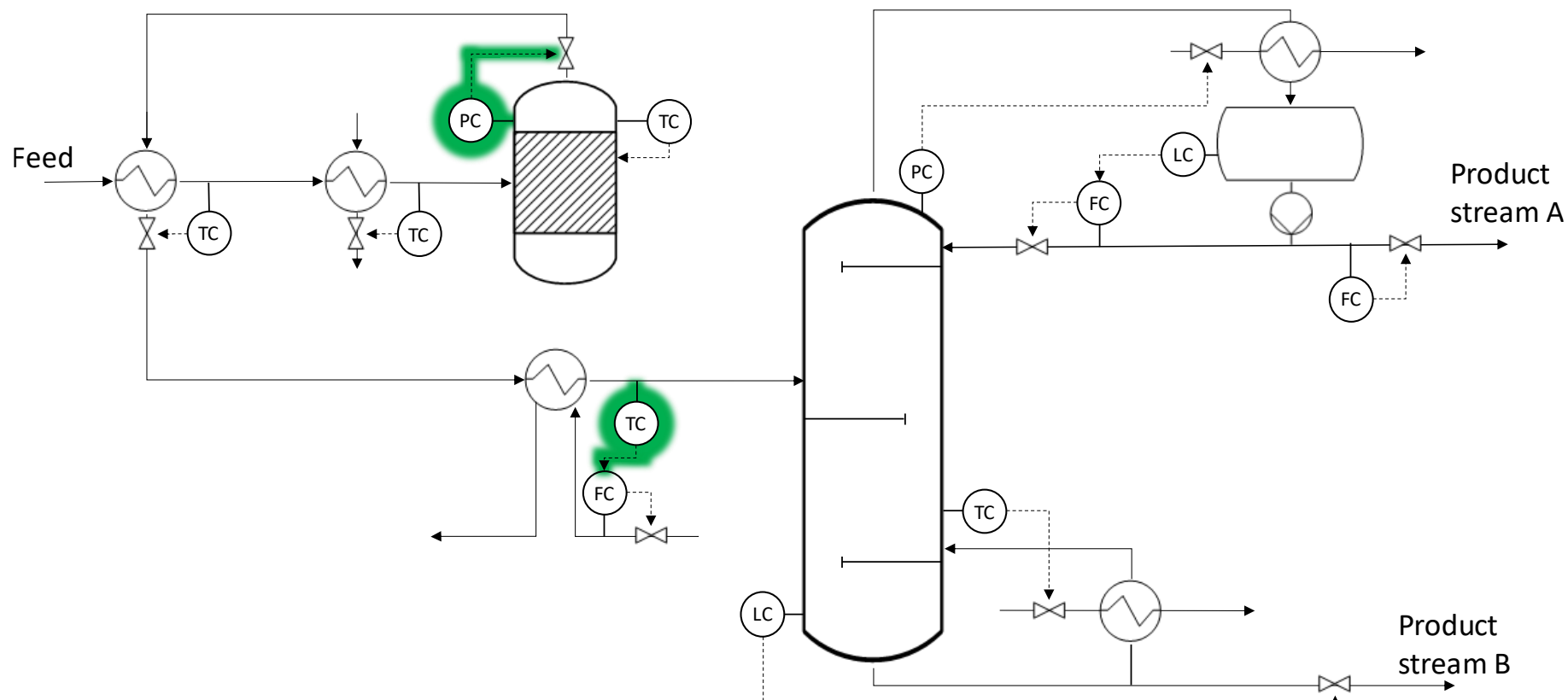

Figure 2. Illustrative case study generated from flowsheet patterns. In the erroneous flowsheet, the pressure and temperature controller highlighted in green are missing. The autocorrection model correctly suggests adding the missing controllers.

While the majority of the model predictions are correct, there are also several false predictions. In case the output flowsheet from the model is incorrect, we observe two common mistakes. Firstly, the model does not or only partially corrects the existing error in the flowsheet. Secondly, the model introduces new errors in the flowsheet. In some cases, the model also alters the SFILES such that the SFILES 2.0 notation is violated and the flowsheet cannot be processed.

Overall, our results suggest that our proposed autocorrection model can learn to autocorrect flowsheets. In particular, the model can add missing components/connections, remove components/connections, and even rearrange components which is a significant advantage over previous works. Also, the LLM has a high learning capacity which has already proven to be effective for autocorrection of human text. While our initial proof of concept shows promising results, there are still several shortcomings and scientific challenges that should be addressed in future research. First, the current model is limited to the topology information of the flowsheet. In the future, further information should be added to the model. For example, (knowledge or hyper) graph representations of flowsheets and graph neural networks are a promising avenue for future research. The graph representation would also allow to increase the complexity of the diagrams (e.g., considering complex industry P&IDs). Second, our current model is trained on synthetic data limiting its industrial application and relevance. In the future, we envision to establish an industry-relevant training dataset. Also, further rules could improve the quality of the synthetic data. Third, various other model architectures such as other LLMs, diffusion models, (variational) autoencoders, or graph-to-sequence could be explored. Finally, the current approach lacks physical/engineering knowledge. We envision to integrate such knowledge into future autocorrection models. For instance, rule-based methods can be integrated with data-driven approaches.

## 5. Conclusions

We present a novel methodology for the autocorrection of flowsheets. Our methodology uses SFILES to represent flowsheets as strings and implements LLMs to automatically correct errors in chemical process flowsheets. Our results show that the proposed autocorrection model can successfully correct flowsheet topologies based on a synthetic dataset with a (top-1) accuracy of 80.1%.

We envision that our autocorrection methodology will become a standard tool in chemical process engineering much like text autocorrection in Word. Our method has the potential to give engineers direct feedback on their process design. By accepting or declining suggestions, the engineer can improve the process design. At the same time, new training data can be generated for the algorithm. Moreover, the methodology of autocorrection can be transferred to other disciplines including technical drawings in mechanical, civil, or electrical engineering.